\documentclass[10pt,twocolumn,letterpaper]{article}

\usepackage{3dv}
\usepackage{times}
\usepackage{epsfig}
\usepackage{graphicx}
\usepackage{amsmath}
\usepackage{amssymb}

\makeatletter
\@namedef{ver@everyshi.sty}{}
\makeatother

\usepackage{float}
\usepackage{color}
\usepackage{tikz}
\usepackage{graphics}
\usepackage{subcaption}
\usepackage{calc}
\usepackage{booktabs}
\definecolor{ToRephaseColor}{rgb}{1.0, 0.4, 0.0}
\definecolor{TodoColor}{rgb}{1.0, 0.0, 1.0}
\definecolor{GreenColor}{rgb}{0.0, 0.0, 1.0}
\definecolor{PurpleColor}{rgb}{0.5, 0.0, 0.5}
\definecolor{BlueColor}{rgb}{0.0, 0.0, 1.0}
\definecolor{BabyBlueColor}{rgb}{0.5, 0.5, 1}
\definecolor{RevisionFixedColor}{rgb}{0.6, 0.0, 0.4}
\definecolor{FinalColor}{rgb}{0.0, 0.0, 0.0}

\def\scheckmark{\tikz\fill[scale=0.1](-0.75, 1) -- (-1.25, 0.5) -- (0, -0.75) -- (2.75, 1.75) -- (2.25, 2.25) -- (0, 0.25) -- cycle;} 
\def\checkmark{\resizebox{\widthof{\scheckmark}*\ratio{\widthof{x}}{\widthof{\normalsize x}}}{!}{\scheckmark}}

\def\scrossmark{\tikz\fill[scale=0.08](-1.5, 2) -- (-2, 1.5) -- (-0.5, 0) -- (-2, -1.5) -- (-1.5, -2) -- (0, -0.5) -- (1.5, -2) -- (2, -1.5) -- (0.5, 0) -- (2, 1.5) -- (1.5, 2) -- (0, 0.5) -- cycle;} 
\def\crossmark{\resizebox{\widthof{\scrossmark}*\ratio{\widthof{x}}{\widthof{\normalsize x}}}{!}{\scrossmark}}

\def\Vhrulefill{\leavevmode\leaders\hrule height 0.7ex depth \dimexpr0.4pt-0.7ex\hfill\kern0pt}

\newcommand{\shortcite}{\cite}

\usepackage[pagebackref=true,breaklinks=true,colorlinks,bookmarks=false]{hyperref}

\threedvfinalcopy %

\def\threedvPaperID{239} %

\ifthreedvfinal\fi
\begin{document}

\title{Progressive Multi-scale Light Field Networks}

\author{David Li\thanks{Email: dli7319@umd.edu\newline
Project page: \url{https://augmentariumlab.github.io/multiscale-lfn/}} \qquad Amitabh Varshney\\
University of Maryland, College Park
}

\maketitle

\begin{abstract}
Neural representations have shown great promise in their ability to represent radiance and light fields while being very compact compared to the image set representation. However, current representations are not well suited for streaming as decoding can only be done at a single level of detail and requires downloading the entire neural network model. Furthermore, high-resolution light field networks can exhibit flickering and aliasing as neural networks are sampled without appropriate filtering. To resolve these issues, we present a progressive multi-scale light field network that encodes a light field with multiple levels of detail. Lower levels of detail are encoded using fewer neural network weights enabling progressive streaming and reducing rendering time. Our progressive multi-scale light field network addresses aliasing by encoding smaller anti-aliased representations at its lower levels of detail. Additionally, per-pixel level of detail enables our representation to support dithered transitions and foveated rendering.
\end{abstract}

\section{Introduction}

Volumetric images and video allow users to view 3D scenes from novel viewpoints with a full 6 degrees of freedom (6DOF). Compared to conventional 2D and 360 images and videos, volumetric content enables additional immersion with motion parallax and binocular stereo while also allowing users to freely explore the full scene.

Traditionally, scenes generated with 3D modeling have been commonly represented as meshes and materials. These classical representations are suitable for real-time rendering as well as streaming for 3D games and AR effects. However, real captured scenes are challenging to represent using the mesh and material representation, requiring complex reconstruction and texture fusion techniques \cite{duo2016fusion4d,du2018montage4d}. To produce better photo-realistic renders, existing view-synthesis techniques have attempted to adapt these representations with multi-plane images (MPI)~\cite{zhou2018stereo}, layered-depth images (LDI)~\cite{shade1998ldi}, and multi-sphere images (MSI)~\cite{attal2020matryodshka}. These representations allow captured scenes to be accurately represented with 6DOF but only within a limited area.

\begin{figure}[!tbp]
    \includegraphics[width=\linewidth]{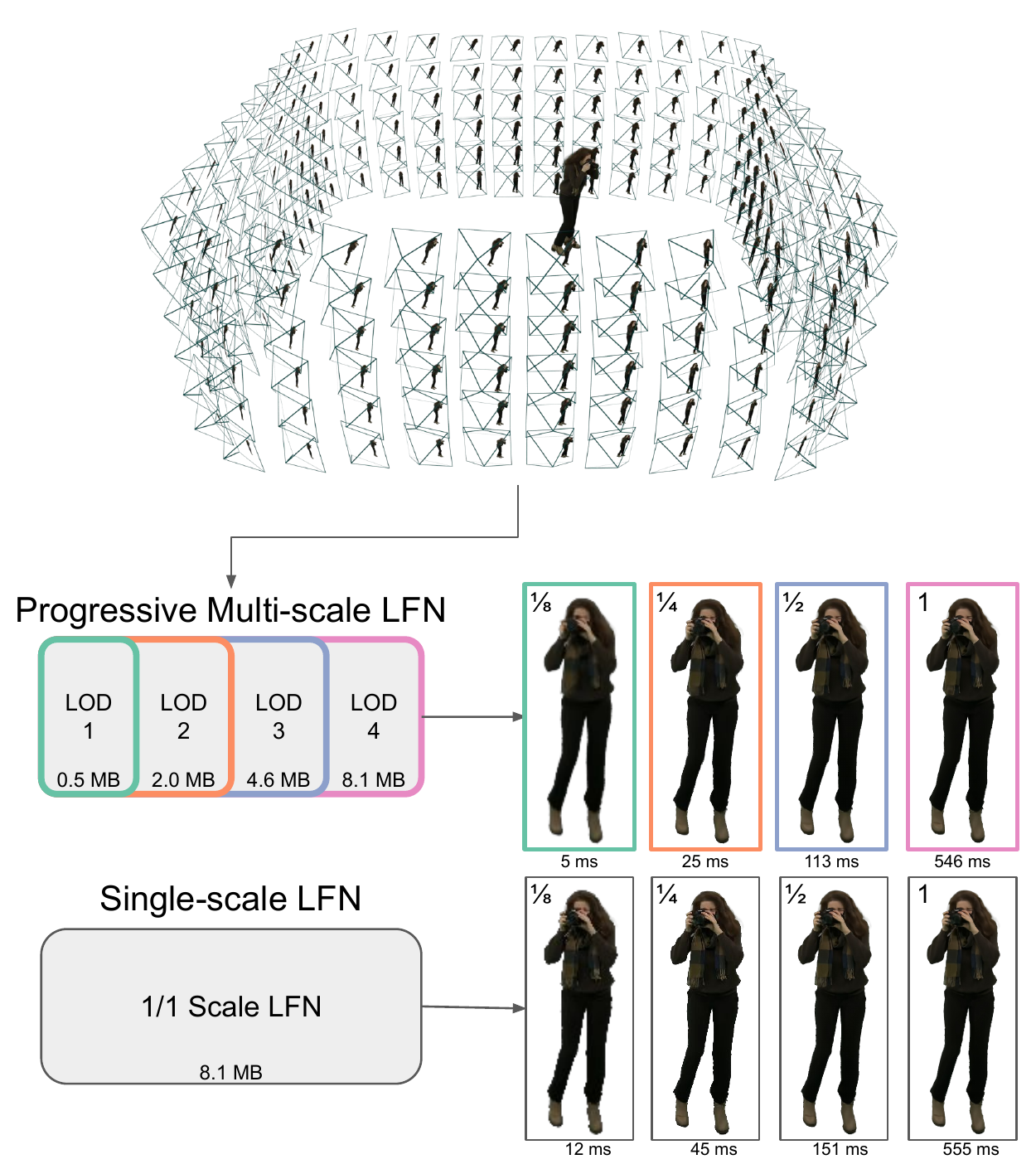}
    \caption{Our progressive multi-scale light field network is suited for streaming, reduces aliasing and flicking, and renders more efficiently at lower resolutions.}
    \label{fig:teaser}
    \vspace{-5mm}
\end{figure}

Radiance fields and light fields realistically represent captured scenes from an arbitrarily large set of viewpoints.
Recently, neural representations such as neural radiance fields (NeRFs)~\cite{mildenhall2020nerf} and light field networks (LFNs)~\cite{sitzmann2021lfns} have been found to compactly represent 3D scenes and perform view-synthesis allowing re-rendering from arbitrary viewpoints. Given a dozen or more photographs capturing a scene from different viewpoints from a conventional camera, a NeRF or LFN can be optimized to accurately reproduce the original images and stored in less than 10 MB. Once trained, NeRFs will also produce images from different viewpoints with photo-realistic quality.

While NeRFs are able to encode and reproduce real-life 3D scenes from arbitrary viewpoints with photorealistic accuracy, they suffer from several drawbacks which limit their use in a full on-demand video streaming pipeline. Some notable drawbacks include slow rendering, aliasing at smaller scales, and a lack of progressive decoding. While some of these drawbacks have been independently addressed in follow-up work~\cite{barron2021mipnerf,yu2021plenoctrees}, approaches to resolving them cannot always be easily combined and may also introduce additional limitations.

In this paper, we extend light field networks with multiple levels of detail and anti-aliasing at lower scales making them more suitable for on-demand streaming. This is achieved by encoding multiple reduced-resolution representations into a single network. With multiple levels of details, light field networks can be progressively streamed and rendered based on the desired scale or resource limitations. For free-viewpoint rendering and dynamic content, anti-aliasing is essential to reducing flickering when rendering at smaller scales. In summary, the contributions of our progressive multi-scale light field network are:
\begin{enumerate}
\item Composing multiple levels of detail within a single network takes up less space than storing all the levels independently.
\item The progressive nature of our model requires us to only download the parameters necessary up to the desired level of detail. 
\item Our model allows rendering among multiple levels of detail simultaneously. This makes for a smooth transition between multiple levels of detail as well as spatially adaptive rendering (including foveated rendering) without requiring independent models at multiple levels of detail on the GPU.
\item Our model allows for faster inference of lower levels of detail by using fewer parameters.
\item We are able to encode light fields at multiple scales to reduce aliasing when rendering at lower resolutions.
\end{enumerate}
\begin{figure}
    \centering
    \captionsetup[subfigure]{labelformat=empty}
    \newcommand{\qDownsamplingWidth}{0.33\linewidth}
    \begin{subfigure}{\qDownsamplingWidth}
    \captionsetup{justification=centering}
    \centering
    \includegraphics[width=0.8\linewidth]{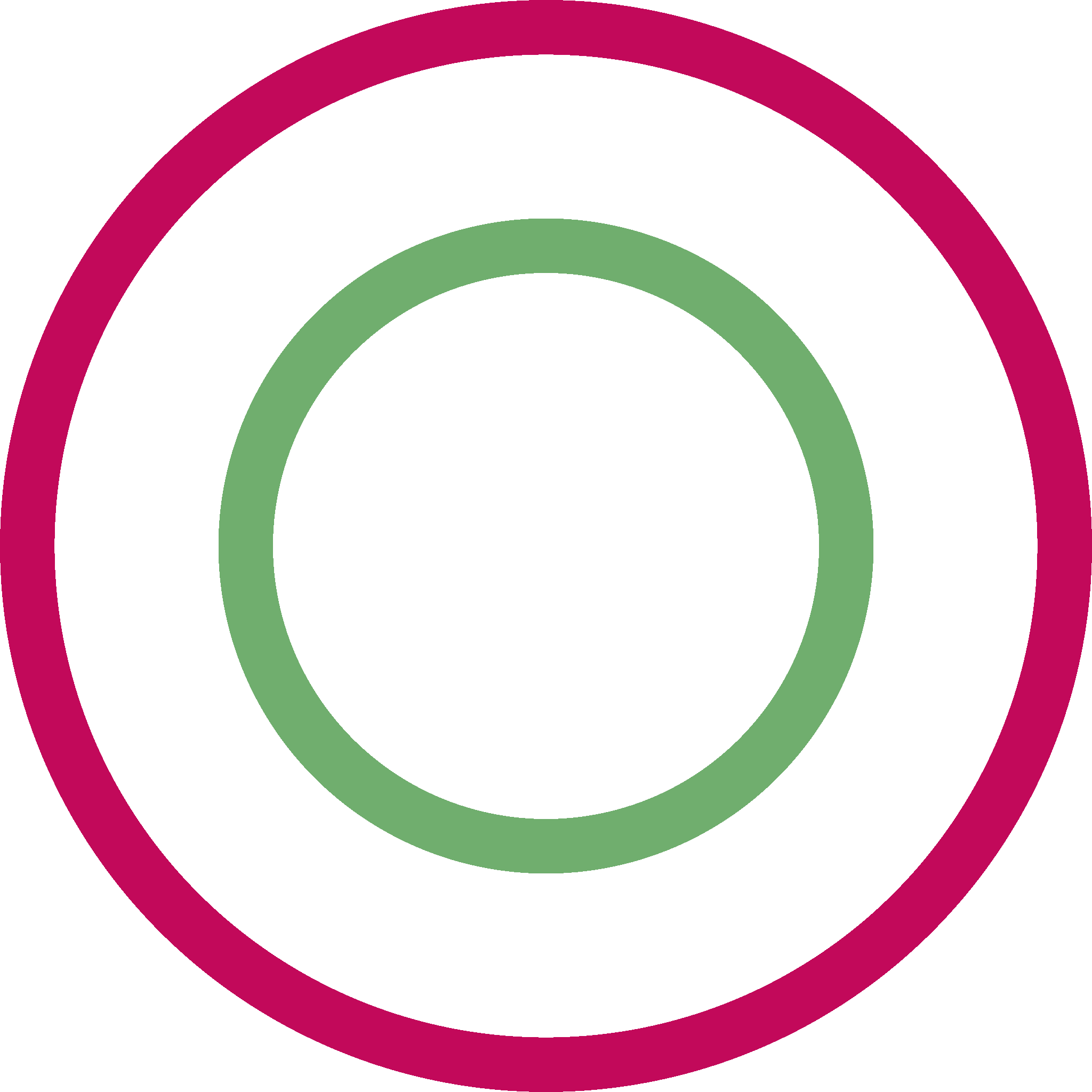}
    \caption{High-resolution\\Reference}
    \end{subfigure}%
    \hfill
    \begin{subfigure}{\qDownsamplingWidth}
    \captionsetup{justification=centering}
    \centering
    \includegraphics[width=0.8\linewidth]{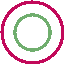}
    \caption{Nearest\\Downsampling}
    \end{subfigure}%
    \hfill
    \begin{subfigure}{\qDownsamplingWidth}
    \captionsetup{justification=centering}
    \centering
    \includegraphics[width=0.8\linewidth]{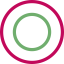}
    \caption{Area\\Downsampling}
    \end{subfigure}%
    \caption{A high-resolution image resized using nearest and area downsampling. Rendering a low-resolution image from a high-resolution light field is similar to performing nearest downsampling, i.e. subsampling a single value. Area downsampling, subsampling with a box filter, generates an anti-aliased image but cannot be done efficiently on an LFN.}
    \label{fig:image_downsampling_methods}
\end{figure}

\section{Background and Related Works}
Our work builds upon existing work in neural 3D representations, adaptive neural network inference, and levels of detail. In this section, we provide some background on neural representations and focus on prior work most related to ours. For a more comprehensive overview of neural rendering, we refer interested readers to a recent survey~\cite{tewari2021advances}.

\begin{table}[!htbp]
  \centering
  \caption{Neural 3D Representations Comparison}
  \label{tab:comparison_table}
  \scalebox{0.80}{
      \begin{tabular}{lcccc}
        \toprule
        Model & Real-time & Model Size & Multi-scale & Progressive\\
        \midrule
        NeRF & \crossmark & $\leq 10$ MB & \crossmark  & \crossmark \\
        mip-NeRF & \crossmark  & $\leq 10$ MB & \checkmark & \crossmark \\
        Plenoctree & \checkmark & $\geq 30$ MB& \crossmark &\crossmark \\
        LFN & \checkmark &  $\leq 10$ MB & \crossmark & \crossmark \\
        Ours & \checkmark & $\leq 10$ MB & \checkmark & \checkmark\\
      \bottomrule
      \end{tabular}
  }
\end{table}

\subsection{Neural 3D Representations}
Traditionally, 3D content has been encoded in a wide variety of formats such as meshes, point clouds, voxels, multi-plane images, and even side-by-side video.
To represent the full range of visual effects from arbitrary viewpoints, researchers have considered using radiance fields and light fields which can be encoded using neural networks. With neural networks, 3D data such as signed distance fields, radiance fields, and light fields can be efficiently encoded as coordinate functions without explicit limitations on the resolution.

\subsubsection{Neural Radiance Field (NeRF)}
Early neural representations focused on signed distance functions and radiance fields. Neural Radiance Fields (NeRF)~\cite{mildenhall2020nerf} use differentiable volume rendering to encode multiple images of a scene into a radiance field encoded as a neural network. With a sufficiently dense set of images, NeRF is able to synthesize new views by using the neural network to interpolate radiance values across different positions and viewing directions. Since the introduction of NeRF, followup works have added support for deformable scenes \cite{pumarola2020d,park2021nerfies,park2021hypernerf}, real-time inference \cite{garbin2021fastnerf,reiser2021kilonerf,neff2021donerf,yu2021plenoxels},
improved quality \cite{shen2021snerf}, generalization across scenes \cite{yu2020pixelnerf,rebain2021lolnerf},
generative modelling \cite{schwarz2020graf,niemeyer2021campari,xie2021fignerf}, videos \cite{li2021neural,xian2021space,wang2022fourier}, sparse views \cite{Chibane_2021_CVPR,niemeyer2021regnerf,Jain_2021_ICCV}, fast training \cite{mueller2022instant,yu2021plenoxels}, and even large-scale scenes \cite{rematas2021urban,xiangli2021citynerf,turki2021meganerf}. 

Among NeRF research, Mip-NeRF~\cite{barron2021mipnerf,barron2022mipnerf360} and BACON~\cite{lindell2021bacon} share a similar goal to our work in offering a multi-scale representation to reduce aliasing. 
Mip-NeRF approximates the radiance across conical regions along the ray, cone-tracing, using integrated positional encoding (IPE). With cone-tracing, Mip-NeRF reduces aliasing while also slightly reducing training time.
BACON~\cite{lindell2021bacon} provide a multi-scale scene representation through band-limited networks using Multiplicative Filter Networks~\cite{fathony2021multiplicative} and multiple exit points. Each scale in BACON has band-limited outputs in Fourier space.

\subsubsection{Light Field Network (LFN)}
Our work builds upon Light Field Networks (LFNs)~\cite{sitzmann2021lfns,feng2021signet,li2022neulf,chandramouli2021light}. Light field networks encode light fields~\cite{levoy1996lightfieldrendering} using a coordinate-wise representation, where for each ray $\mathbf{r}$, a neural network $f$ is used to directly encode the color $c=f(\mathbf{r})$. To represent rays, Feng and Varshney \shortcite{feng2021signet} pair the traditional two-plane parameterization with Gegenbauer polynomials while Sitzmann \etal~\shortcite{sitzmann2021lfns} use Pl\"{u}cker coordinates which combine the ray direction $\mathbf{r}_d$ and ray origin $\mathbf{r}_o$ into a 6-dimensional vector $(\mathbf{r}_d, \mathbf{r}_o \times \mathbf{r}_d)$. Light fields can also be combined with explicit representations~\cite{ost2021point}. In our work, we adopt the Pl\"{u}cker coordinate parameterization which can represent rays in all directions.

Compared to NeRF, LFNs are faster to render as they only require a single forward pass per ray, enabling real-time rendering. However, LFNs are worse at view synthesis as they lack the self-supervision that volume rendering provides with explicit 3D coordinates. As a result, LFNs are best trained from very dense camera arrays or from a NeRF teacher model such as in~\cite{attal2022learning}. Similar to NeRFs, LFNs are encoded as multi-layer perceptrons, hence they remain compact compared to voxel and octree NeRF representations~\cite{yu2021plenoctrees,hedman2021snerg,garbin2021fastnerf,yu2021plenoxels} which use upwards of $50$ MB. Therefore LFNs strike a balance between fast rendering times and storage compactness which could make them suitable for streaming 3D scenes over the internet.

\subsection{Levels of Detail}
Level of detail methods are commonly used in computer graphics to improve performance and quality. For 2D textures, one of the most common techniques is mipmapping~\cite{williams1983mipmap} which both reduces aliasing and improves performance with cache coherency by encoding textures at multiple resolutions. For neural representations, techniques for multiple levels of detail have also been proposed for signed distance functions as well as neural radiance fields.

For signed distance fields, Takikawa \etal \shortcite{takikawa2021neural} propose storing learned features within nodes of an octree to represent a signed distance function with multiple levels of detail. 
Building upon octree features, Chen \etal \shortcite{chen2021multiresolution} develop Multiresolution Deep Implicit Functions (MDIF) to represent 3D shapes which combines a global SDF representation with local residuals. 
For radiance fields, Yang \etal~\shortcite{yang2021recursivenerf} develop Recursive-NeRF which grows a neural representation with multi-stage training which is able to reduce NeRF rendering time. 
BACON~\cite{lindell2021bacon} offers levels of detail for various scene representations including 2D images and radiance fields with different scales encoding different bandwidths in Fourier space.
PINs~\cite{Landgraf2022PINs} uses a progressive fourier encoding to improve reconstruction for scenes with a wide frequency spectrum.
For more explicit representations, Variable Bitrate Neural Fields~\cite{takikawa2022variable} use a vector-quantized auto-decoder to compress feature grids into lookup indices and a learned codebook. Levels of detail are obtained by learning compressed feature grids at multiple resolutions in a sparse octree.
In parallel, Streamable Neural Fields~\cite{cho2022streamable} propose using progressive neural networks~\cite{rusu2016progressive} to represent spectrally, spatially, or temporally growing neural representations.

\subsection{Adaptive Inference}
Current methods use adaptive neural network inference based on available computational resources or the difficulty of each input example. Bolukbasi \etal~\shortcite{bolukbasi2017adaptive} propose two schemes for adaptive inference. Early exiting allows intermediate layers to route outputs directly to an exit head while network selection dynamically routes features to different networks. Both methods allow easy examples to be classified by a subset of neural network layers. Yeo \etal \shortcite{yeo2018neuraladaptive} apply early exiting to upscale streamed video. By routing intermediate CNN features to exit layers, a single super-resolution network can be partially downloaded and applied on a client device based on available compute. Similar ideas have also been used to adapt NLP to input complexity~\cite{zhou2020bert,weijie2020fastbert}.
Yang \etal~\shortcite{yang2020resolutionadaptive} develop Resolution Adaptive Network (RANet) which extracts features and performs classification of an image progressively from low-resolution to high-resolution. Doing so sequentially allows efficient early exiting as many images can be classified with only coarse features. Slimmable neural networks~\cite{yu2018slimmable} train a model at different widths with a switchable batch normalization and observe better performance than individual models at detection and segmentation tasks.

\section{Method}
Our method consists of a multi-scale light field encoded using a progressive neural network to reduce aliasing and enable adaptive streaming and rendering.

\subsection{Multi-scale Light Field}

We propose encoding multiple representations of the light field which produce anti-aliased representations of a light field, similar to mipmaps for conventional images. For this, we generate an image pyramid for each image view with area downsampling and encode each resolution as a separate light field representation. In contrast to bilinear or nearest pixel downsampling, area downsampling creates an anti-aliased view as shown in \autoref{fig:image_downsampling_methods} where each pixel in the downsampled image is an average over a corresponding area in the full-resolution image.

\begin{figure}[!htb]
    \centering
    \captionsetup[subfigure]{labelformat=empty}
    \begin{subfigure}{0.52\linewidth}
        \includegraphics[width=\linewidth]{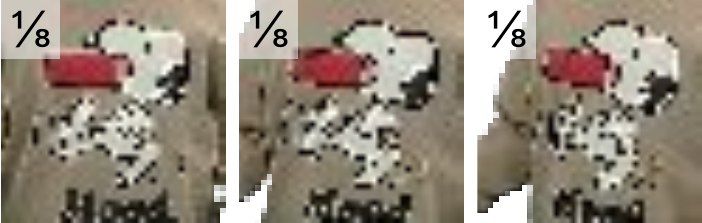}
        \vspace{-5.5mm}
        \caption{Single-scale Light Field Network}
    \end{subfigure}\\
    \vspace{1mm}
    \begin{subfigure}{0.52\linewidth}
        \includegraphics[width=\linewidth]{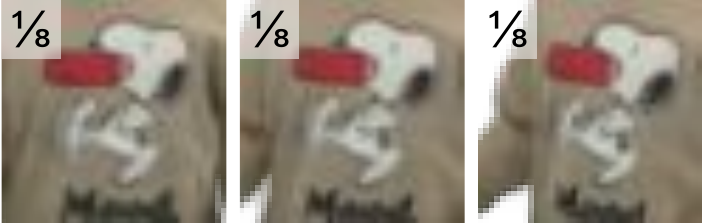}
        \vspace{-5.5mm}
        \caption{Multi-scale Light Field Network}
    \end{subfigure}
    \caption{Rendering a single full-scale LFN at a lower (1/8) resolution results in aliasing, unlike the multi-scale LFN at the same (1/8) resolution. When changing viewpoints (3 shown above), the aliasing results in flickering.}
    \label{fig:aliasing_figure}
\end{figure}

As with mipmaps, lower-resolution representations trade-off sharpness for less aliasing. In this case, where a light field is rendered into an image, sharpness may be preferable over some aliasing. However, in the case of videos where the light field is viewed from different poses or the light field is dynamic, sampling from a high-resolution light field leads to flickering. With a multi-scale representation, rendering light fields at smaller resolutions can be done at a lower scale to reduce flickering.

\subsection{Progressive Model}

\begin{figure}[!htbp]
  \centering
  \includegraphics[width=\linewidth]{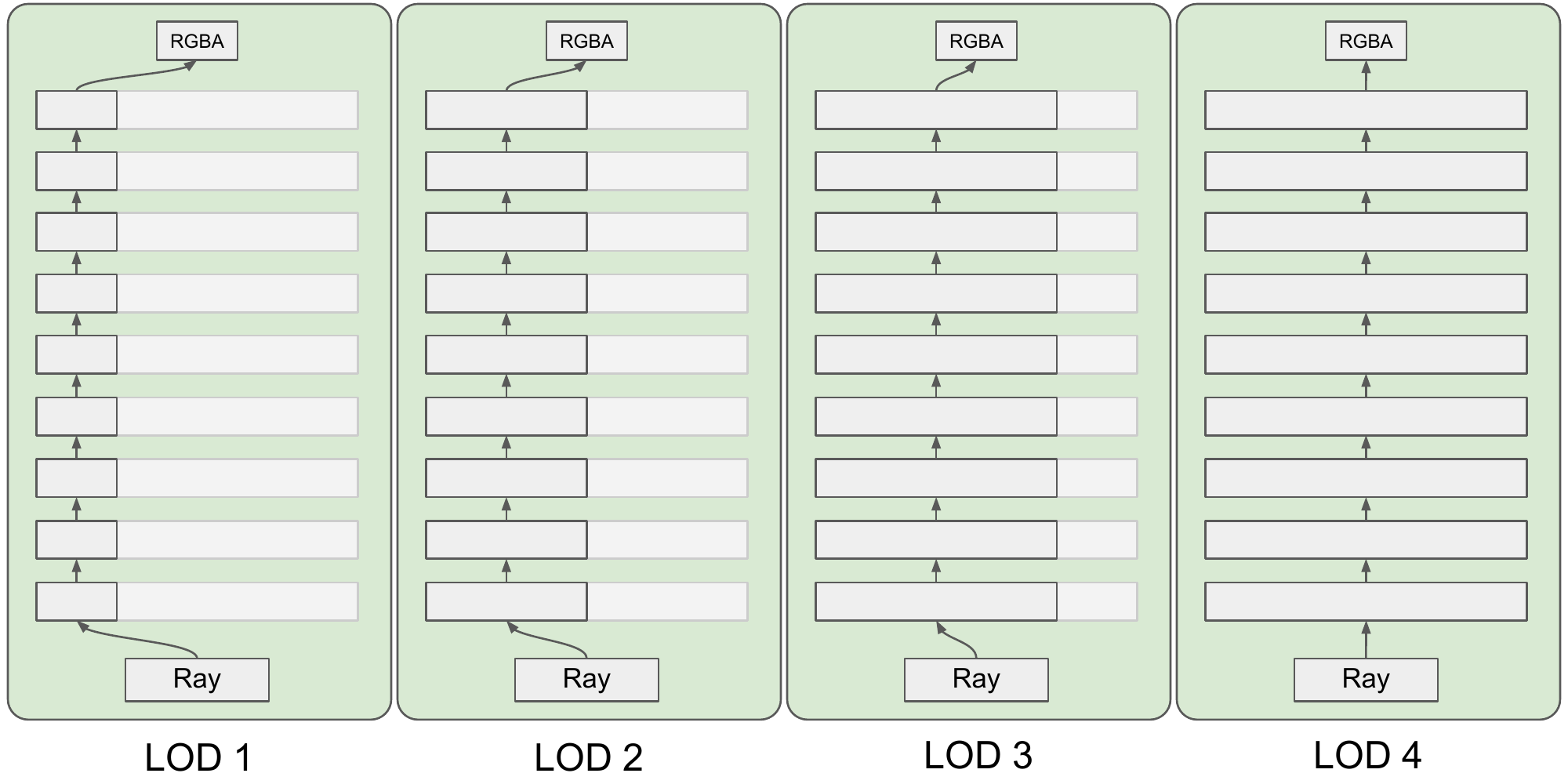}
  \caption{Using subsets of neurons to encode different levels of detail within a single model.}
  \label{fig:model_illustration_subnet}
\end{figure}

For light field streaming, we wish to have a compact representation that can simultaneously encode multiple levels of detail into a single progressive model. Specifically, we seek a model with the following properties:
Given specific subsets of network weights, we should be able to render a complete image with reduced details.
Progressive levels of details should share weights with lower levels to eliminate encoding redundancy.

With the above properties, our model offers several benefits over a standard full-scale network or encoding multi-scale light fields using multiple networks.
First, our model composes all scales within a single network hence requiring less storage space than storing four separate models.
Second, our model is progressive, thus only parameters necessary for the desired level of detail are needed in a streaming scenario.
Third, our model allows rendering different levels of detail across pixels for dithered transitions and foveation.
Fourth, our model allows for faster inference of lower levels of detail by utilizing fewer parameters.
Model sizes and training times appear in \autoref{tab:nonprogressive_vs_progressive}.

\begin{figure}[!htbp]
    \centering
    \includegraphics[width=0.5\linewidth]{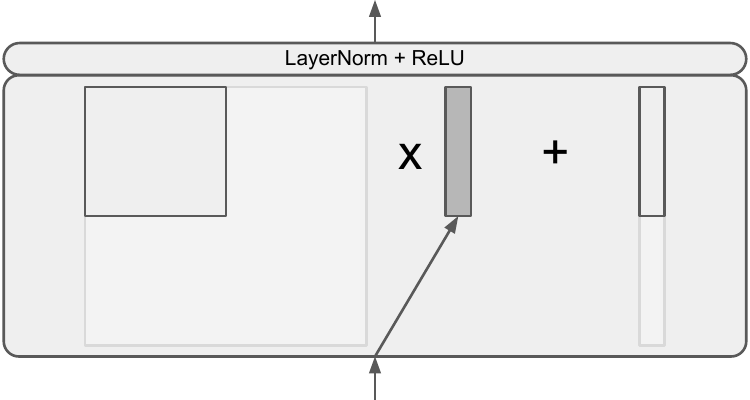}
    \caption{Illustration of a single layer where a subset of the weight matrix and bias vector are used, evaluating half of the neurons and reducing the operations down to approximately a quarter.}
    \label{fig:subnet_layer_illustration}
\end{figure}

\subsubsection{Subset of neurons}
To encode multiple levels of detail within a unified representation, we propose using subsets of neurons as shown in \autoref{fig:model_illustration_subnet}. For each level of detail, we assign a fixed number of neurons to evaluate, with increasing levels of detail using an increasing proportion of the neurons in the neural network. An illustration of the matrix subset operation applied to each layer is shown in \autoref{fig:subnet_layer_illustration}. For example, a single linear layer with $512$ neurons will have a $512 \times 512$ weight matrix and a bias vector of size $512$. For a low level of detail, we can assign a neuron subset size of $25\%$ or $128$ neurons. Then each linear layer would use the top-left $128 \times 128$ submatrix for the weights and the top $128$ subvector for the bias. To keep the input and output dimensions the same, the first hidden layer uses every column of the weight matrix and the final output layer uses every row of the weight matrix.
Unlike using a subset of layers, as is done with early exiting~\cite{bolukbasi2017adaptive,yeo2018neuraladaptive}, using a subset of neurons preserves the number of sequential non-linear activations between all levels of details. Since layer widths are typically larger than model depths, using subsets of neurons also allows more granular control over the quantity and quality of levels. 

\subsubsection{Training} \label{sec:mipnet_training}
With multiple levels of detail at varying resolutions, a modified training scheme is required to efficiently encode all levels of detail together. 
On one hand, generating all levels of detail at each iteration heavily slows down training as each iteration requires an independent forward pass through the network. 
On the other hand, selecting a single level of detail at each iteration or applying progressive training compromises the highest level of detail whose full weights would only get updated a fraction of the time.

We adopt an intermediate approach that focuses on training the highest level of detail. 
During training, each batch does a full forward pass through the entire network, producing pixels at the highest level of detail along with a forward pass at a randomly selected lower level of detail. The squared L2 losses for both the full detail and reduced detail are added together for a combined backward pass. By training at the highest level of detail, we ensure that every weight gets updated at each batch and that the highest level of detail is trained to the same number of iterations as a standard model. Specifically, given a batch of rays $\{\mathbf{r}_i\}_{i=1}^{b}$, corresponding ground truth colors $\{\mathbf\{\mathbf{y}_i^c\}_{i=1}^{b}\}_{c=1}^{4}$ for four levels of details, and a random lower level of detail $k \in \{1,2,3\}$, our loss function is:
$$L = \frac{1}{b}\sum_{i=1}^b \left[ \left\Vert f_4(\mathbf{r}_i) - \mathbf{y}_i^4\right\Vert^2_2 + \left\Vert f_k(\mathbf{r}_i) - \mathbf{y}_i^k\right\Vert^2_2 \right]$$
As rays are sampled from the highest-resolution representation, each ray will not correspond to a unique pixel in the lower-resolution images in the image pyramid. Thus for lower levels of detail, corresponding colors are sampled using bilinear interpolation from the area downsampled training images.

\subsection{Adaptive Rendering}
Our proposed model allows dynamic levels of detail on a per-ray/per-pixel level or on a per-object level. As aliasing appears primarily at smaller scales, selecting the level of detail based on the distance to the viewer reduces aliasing and flickering for distant objects.

\subsubsection{Distance-based Level of Detail}
When rendering light fields from different distances, the space between rays in object space is proportional to the distance between the object and the camera. Due to this, aliasing and flickering artifacts can occur at large distances when the object appears small. By selecting the level of detail based on the distance of the object from the camera or viewer, aliasing and flickering can be reduced. With fewer pixels to render, rendering smaller objects involves fewer forward passes through the light field network. Hence, the amount of operations is typically proportional to the number of pixels. By rendering smaller representations at lower LODs, we further improve the performance as pixels from lower LODs only perform a forward pass using a subset of weights.

\begin{figure}[!htbp]
    \centering
    \includegraphics[width=\linewidth]{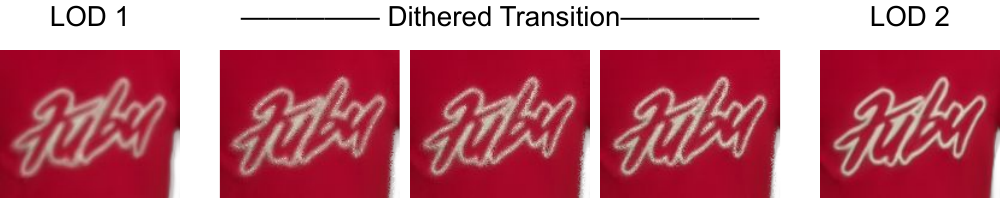}
    \caption{With per-pixel level of detail, dithering can be applied to transition between levels of detail.}
    \label{fig:transition_example}
\end{figure}

\subsubsection{Dithered Transitions}
With per-pixel LOD, transitioning between levels can be achieved with dithered rendering. By increasing the fraction of pixels rendered at the new level of detail in each frame, dithering creates the appearance of a fading transition between levels as shown in \autoref{fig:transition_example}.

\begin{figure}[!htbp]
    \centering
    \includegraphics[width=0.8\linewidth]{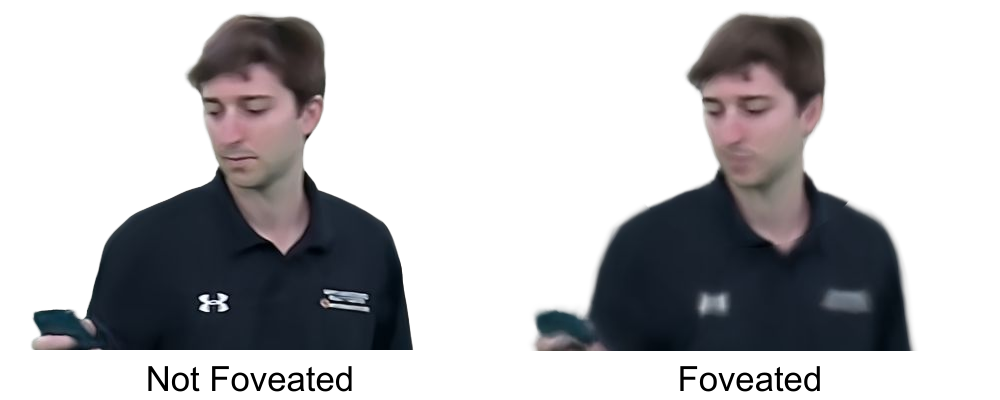}
    \caption{An example of foveated rendering from our progressive multi-scale LFN. Foveated rendering  generates peripheral pixels at lower levels of detail to further improve performance in eye-tracked environments. In this example, foveation is centered over the left eye of the subject. }
    \label{fig:foveation_example}
\end{figure}

\subsubsection{Foveated Rendering}
In virtual and augmented reality applications where real-time gaze information is available through an eye-tracker, foveated rendering~\cite{Meng20203D,meng2018Kernel,li2021logrectilinear} can be applied to better focus rendering compute. As each pixel can be rendered at a separate level of detail, peripheral pixels can be rendered at a lower level of detail to improve the performance. \autoref{fig:foveation_example} shows an example of foveation from our progressive multi-scale LFN.

\newcommand{\qQualitativeWidth}{0.24\linewidth}
\newcommand{\addDatasetResults}[2]{
    \begin{subfigure}[b]{\qQualitativeWidth}
        \centering
        \includegraphics[width=0.8\linewidth]{figures/#2/comparison2_r8.pdf}
        \caption{Dataset #1 at 1/8 scale}
        \label{fig:qualitative_comparison_#1_r8}
    \end{subfigure}
    \hfill
    \begin{subfigure}[b]{\qQualitativeWidth}
        \centering
        \includegraphics[width=0.8\linewidth]{figures/#2/comparison2_r4.pdf}
        \caption{Dataset #1 at 1/4 scale}
        \label{fig:qualitative_comparison_#1_r4}
    \end{subfigure}
    \hfill
    \begin{subfigure}[b]{\qQualitativeWidth}
        \centering
        \includegraphics[width=0.8\linewidth]{figures/#2/comparison2_r2.pdf}
        \caption{Dataset #1 at 1/2 scale}
        \label{fig:qualitative_comparison_#1_r2}
    \end{subfigure}
    \hfill
    \begin{subfigure}[b]{\qQualitativeWidth}
        \centering
        \includegraphics[width=0.8\linewidth]{figures/#2/comparison2_r1.pdf}
        \caption{Dataset #1 at full scale}
        \label{fig:qualitative_comparison_#1_r1}
    \end{subfigure}
}
\begin{figure*}[!htbp]
    \centering
    \addDatasetResults{C}{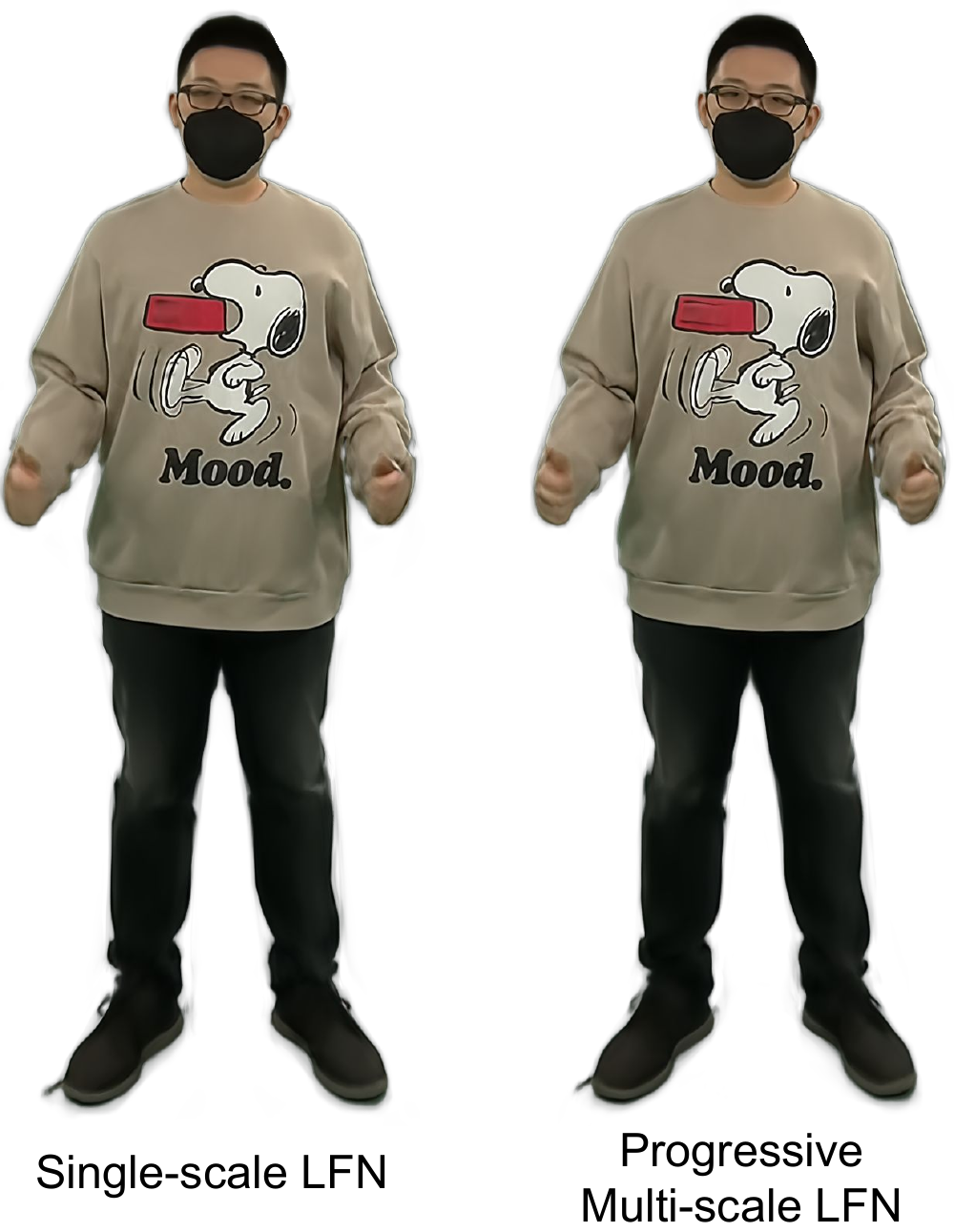}
    \caption{Qualitative results of one of our datasets at four scales. Single-scale LFNs (left) trained at the full-resolution exhibit aliasing and flickering at lower scales. Our progressive multi-scale LFNs (right) encode all four scales into a single model and have reduced aliasing and flickering at lower scales.}
    \label{fig:qualitative_comparison}
\end{figure*}

\newcommand{\datasetScaledResults}[2]{
    \begin{subfigure}[b]{0.17\linewidth}
      \centering
      \includegraphics[width=\linewidth]{figures/#2/mipnet_lod_scaled.pdf}
      \caption{Dataset #1}
      \label{fig:qualitative_scaled_#2}
    \end{subfigure}
}
\begin{figure*}[!htbp]
    \centering
    \datasetScaledResults{A}{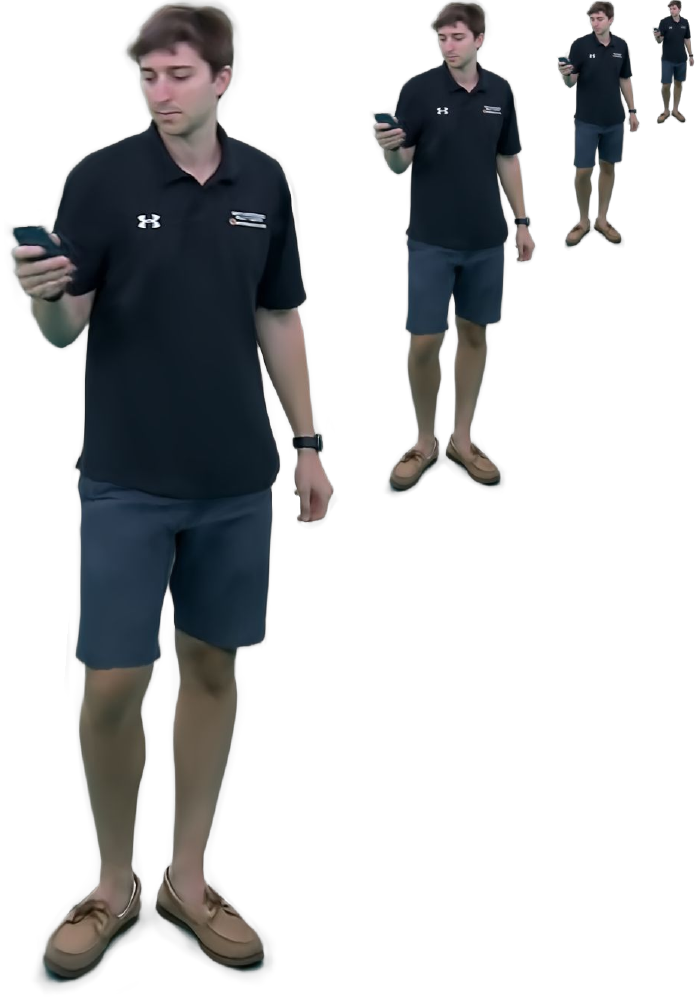}
    \hfill
    \datasetScaledResults{B}{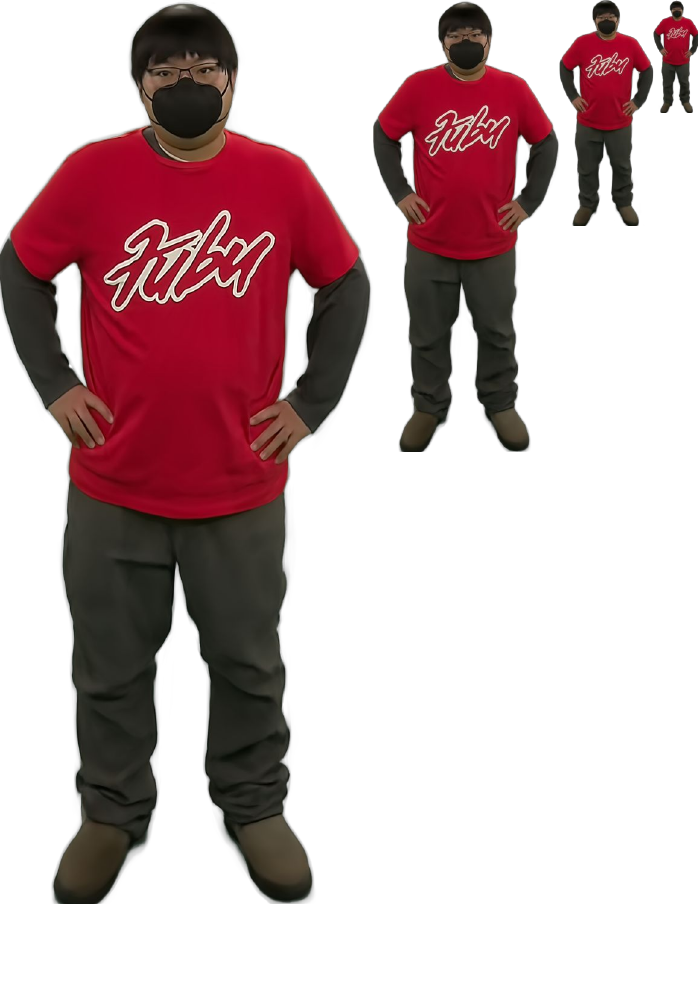}
    \hfill
    \datasetScaledResults{C}{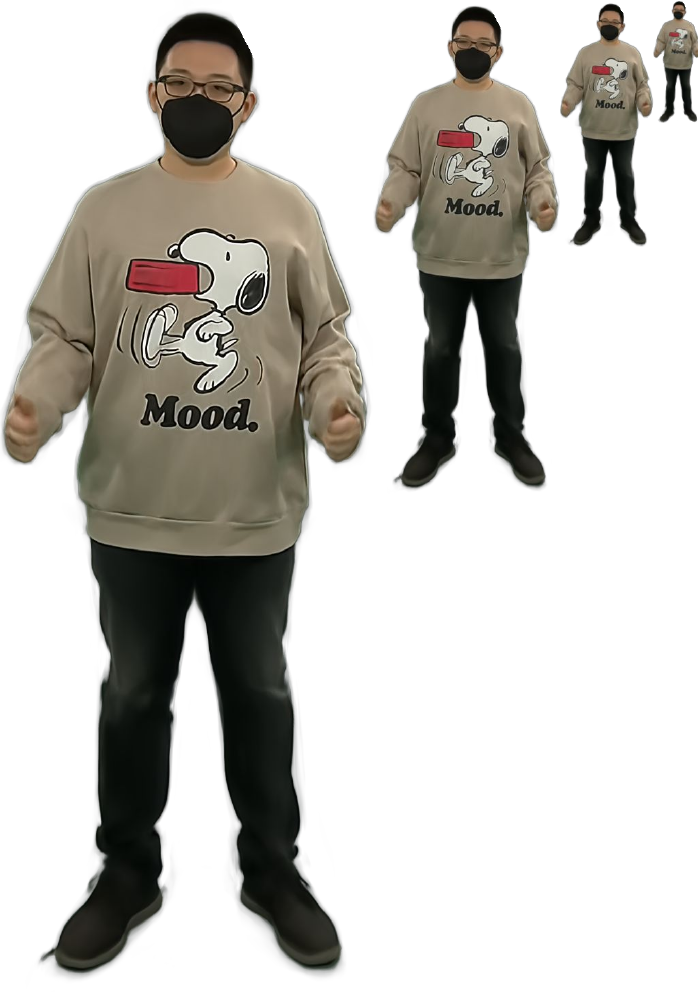}
    \hfill
    \datasetScaledResults{D}{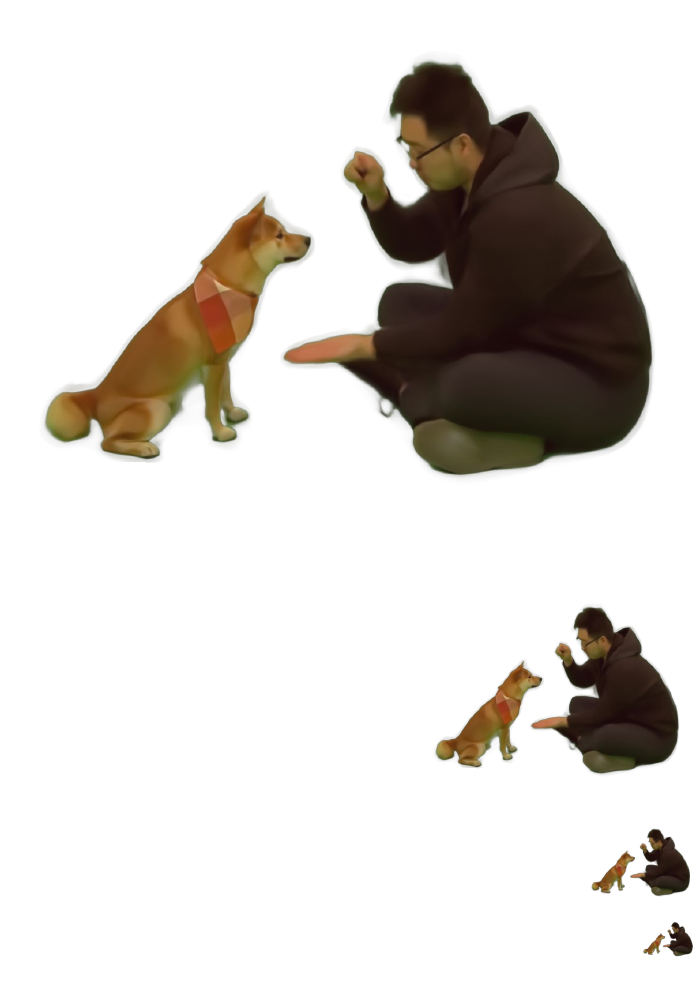}
    \hfill
    \datasetScaledResults{E}{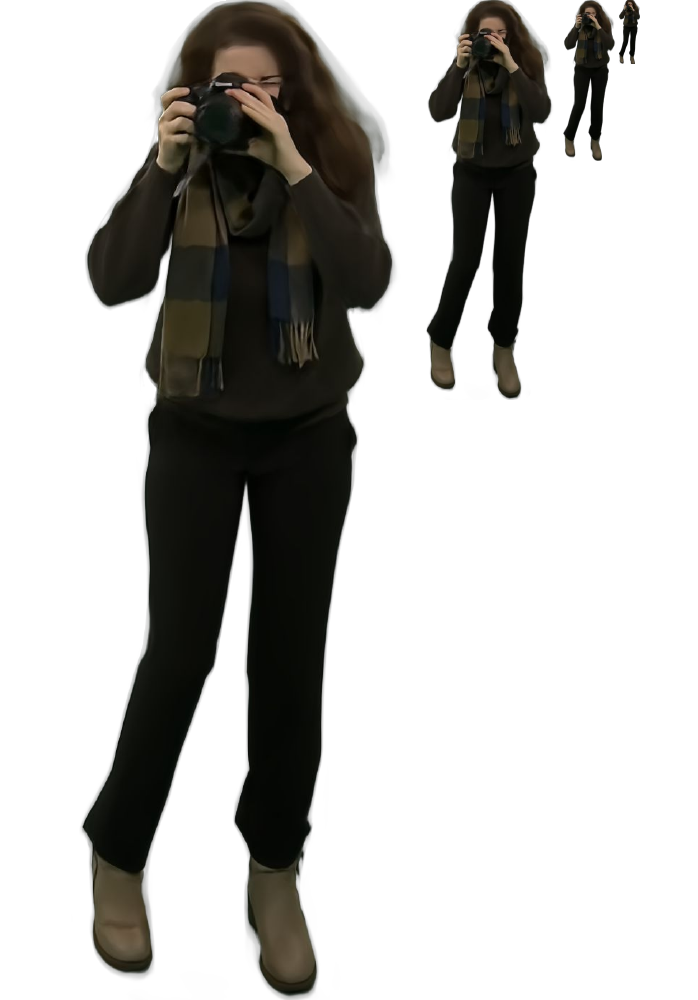}
    \caption{Scaled qualitative results from our progressive multi-scale LFNs at four levels of detail.}
    \label{fig:qualtiative_scaled}
\end{figure*}

\section{Experiments}
To evaluate our progressive multi-scale LFN, we conduct experiments with four levels of detail. We first compare the quality when rendering at different scales from a standard single-scale LFN and our multi-scale LFN. We then perform an ablation comparing multiple LFNs trained at separate scales to our progressive network. Finally, we benchmark our multi-scale LFN to evaluate the speedup gained by utilizing fewer neurons when rendering at lower levels of detail. Additional details are in the supplementary material.

\subsection{Experimental Setup}
For our evaluation, we use datasets of real people consisting of $240$ synchronized images from our capture studio. Images are captured at $4032 \times 3040$ resolution. Our datasets are processed with COLMAP~\cite{schoenberger2016sfm,schoenberger2016mvs} to extract camera parameters and background matting~\cite{lin2020matting} to focus on the foreground subject. In each dataset, images are split into groups of 216 training poses, 12 validation poses, and 12 test poses.

We quantitatively evaluate the encoding quality by computing the PSNR and SSIM metrics. Both metrics are computed on cropped images that tightly bound the subject to reduce the contribution of empty background pixels. We also measure the render time to determine the relative speedup at different levels of detail. Metrics are averaged across all images in the test split of each dataset. 

When training LFNs, we use a batch size of $8192$ rays with $67\%$ of rays sampled from the foreground and $33\%$ sampled from the background. For each epoch, we iterate through training images in batches of two images. In each image batch, we train on $50\%$ of all foreground rays sampling rays across the two viewpoints. For our multi-scale progressive model and single-scale model, we train using $100$ epochs. For our ablation LFNs at lower resolutions, we train until the validation PSNR matches that of our progressive multi-scale LFN with a limit of $1000$ epochs. Each LFN is trained using a single NVIDIA RTX 2080 TI GPU.

\begin{table}[!htbp]
  \caption{Model Parameters at Different Levels of Detail.}
  \label{tab:model_parameters}
  \centering
  \scalebox{0.87}{
  \begin{tabular}{lrrrr}
    \toprule
    Level of Detail & 1 & 2 & 3 & 4\\
    \midrule
    Model Layers & 10 & 10 & 10 & 10\\
    Layer Width  & 128 & 256 & 384 & 512\\
    Parameters   & 136,812 & 533,764 & 1,193,860 & 2,116,100\\
    Full Size (MB)    & 0.518 & 2.036 & 4.554 & 8.072\\
    Target Scale    & $1/8$ & $1/4$ & $1/2$ & $1$\\
    Train Width & 504 & 1008 & 2016 & 4032\\
    Train Height & 380 & 760 & 1520 & 3040\\
    \bottomrule
  \end{tabular}
  }
\end{table}

\subsection{Rendering Quality}
We first evaluate the rendering quality by comparing a standard single-scale LFN trained at the highest resolution with our multi-scale progressive LFN. 
For the single-scale LFN, we train a model with $10$ layers and $512$ neurons per hidden layer on each of our datasets and render them at multiple scales: $1/8$, $1/4$, $1/2$, and full scale. 
For our multi-scale progressive LFN, we train an equivalently sized model with four LODs corresponding to each of the target scales. The lowest LOD targets the $1/8$ scale and utilizes $128$ neurons from each hidden layer. Each increasing LOD uses an additional $128$ neurons. Model parameters for each LOD are laid out in \autoref{tab:model_parameters}.
Note that our qualitative results are cropped to the subject and hence have a smaller resolution.
Qualitative results are shown in \autoref{fig:qualitative_comparison} with renders from both models placed side-by-side. 
Quantitative results are shown in \autoref{tab:quantiative_quality}.

At the full scale, both single-scale and multi-scale models are trained to produce full-resolution images. As a result, the quality of the renders from both models is visually similar as shown in \autoref{fig:qualitative_comparison}.
At $1/2$ scale, the resolution is still sufficiently high so little to no aliasing can be seen in renders from either model. At $1/4$ scale, we begin to see significant aliasing around figure outlines and shape borders in the single-scale model.
This is especially evident in \autoref{fig:qualitative_comparison_C_r4} where the outlines in the cartoon graphic have an inconsistent thickness. 
Our multi-scale model has much less aliasing with the trade-off of having more blurriness. At the lowest level of detail, we see severe aliasing along the fine textures as well as along the borders of the object in the single-scale model. With a moving camera, the aliasing seen at the two lowest scales appears as flickering artifacts.

\begin{table}[!htbp]
  \caption{Average Rendering Quality Over All Datasets.}
  \newcommand{\quantitativeQualityTabScale}{0.9}
  \label{tab:quantiative_quality}
  \centering
  \begin{subfigure}[h]{\linewidth}
      \centering
      \scalebox{\quantitativeQualityTabScale}{
      \begin{tabular}{lcccc}
          \toprule
          Model & LOD 1 & LOD 2 & LOD 3 & LOD 4 \\ 
          \midrule
Single-scale LFN & 26.95 & 28.05 & 28.21 & 27.75 \\
Multiple LFNs & 29.13 & 29.88 & 29.27 & 27.75 \\
Multi-scale LFN & 29.37 & 29.88 & 29.01 & 28.12 \\
          \bottomrule
      \end{tabular}
      }
      \caption{PSNR (dB) at 1/8, 1/4, 1/2, and 1/1 scale.}
      \label{fig:quantiative_quality_psnr}
  \end{subfigure}\\
  \vspace{2mm}
  \begin{subfigure}[h]{\linewidth}
      \centering
      \scalebox{\quantitativeQualityTabScale}{
      \begin{tabular}{lcccc}
          \toprule
           Model & LOD 1 & LOD 2 & LOD 3 & LOD 4 \\ 
          \midrule
Single-scale LFN & 0.8584 & 0.8662 & 0.8527 & 0.8480 \\
Multiple LFNs & 0.8133 & 0.8572 & 0.8532 & 0.8480 \\
Multi-scale LFN & 0.8834 & 0.8819 & 0.8626 & 0.8570 \\
           \bottomrule
      \end{tabular}
      }
      \caption{SSIM at 1/8, 1/4, 1/2, and 1/1 scale.}
      \label{fig:quantiative_quality_ssim}
  \end{subfigure}
\end{table}

\subsection{Progressive Model Ablation}
We next perform an ablation experiment to determine the benefits and drawbacks of our progressive representation.
For a non-progressive comparison, we represent each scale of a multi-scale light field using a separate LFN. 
Each LFN is trained using images downscaled to the appropriate resolution.
This ablation helps determine the training overhead our progressive LFNs encounter by compressing four resolutions into a single model and how the rendering quality compares to having separate models.

\begin{table}[!htbp]
  \caption{Multiple LFNs \textit{vs} Progressive Multi-scale LFN.}
  \newcommand{\nonprogressiveProgressiveTabScale}{0.83}
  \label{tab:nonprogressive_vs_progressive}
  \centering
  \begin{subfigure}[t]{\linewidth}
  \centering
  \scalebox{\nonprogressiveProgressiveTabScale}{
  \begin{tabular}{lccccc}
    \toprule
    Model & LOD 1 & LOD 2 & LOD 3 & LOD 4 & Total\\
    \midrule
    Multiple LFNs & 0.518 & 2.036 & 4.554 & 8.072 & 15.180\\
    Multi-scale LFN & \multicolumn{4}{c}{\Vhrulefill\;8.072\;\Vhrulefill} & 8.072\\
    \bottomrule
  \end{tabular}
  }
  \caption{Model Size (MB).}
  \label{tab:nonprogressive_vs_progressive_model_size}
  \end{subfigure}\\
  \vspace{2mm}
  \begin{subfigure}[t]{\linewidth}
  \centering
  \scalebox{\nonprogressiveProgressiveTabScale}{
  \begin{tabular}{lccccc}
    \toprule
    Model & LOD 1 & LOD 2 & LOD 3 & LOD 4 & Total\\
    \midrule
Multiple LFNs & 3.51 & 6.36 & 4.09 & 11.35 & 25.31\\
Multi-scale LFN & \multicolumn{4}{c}{\Vhrulefill\;17.78\;\Vhrulefill} & 17.78\\
    \bottomrule
  \end{tabular}
  }
  \caption{Average Training Time Over All Datasets (hours).}
  \label{tab:nonprogressive_vs_progressive_training_time}
  \end{subfigure}
\end{table}

Progressive and non-progressive model sizes and training times are shown in \autoref{tab:nonprogressive_vs_progressive}. In terms of model size, using multiple LFNs adds up to $15$ MB compared to $8$ MB for our progressive model. This $47\%$ savings could allow storing or streaming of more light fields where model size is a concern. 
For the training time, we observe that training a multi-scale network takes on average $17.78$ hours. Additional offline training time for our multi-scale network compared to training a single standard LFN is expected since each training iteration involves two forward passes. Encoding a multi-scale light field using multiple independent LFNs requires training each network separately, totaling $25.31$ hours. Despite the higher compression achieved by our progressive multi-scale LFN, we observe little to no training overhead. On average, training our multi-scale progressive LFN saves $30\%$ of training time compared to naively training multiple light field networks at different resolutions.

Quantitative PSNR and SSIM quality metrics are shown in \autoref{tab:quantiative_quality}. We observe that utilizing multiple LFNs to encode each scale of each light field yields better PSNR results compared to rendering from a single-scale LFN. However, utilizing multiple LFNs increases the total model size.
Our multi-scale LFN achieves the superior PSNR and SSIM results between these two setups without increasing the total model size.
In an on-demand streaming scenario, using only a single-scale model would incur visual artifacts and unnecessary computation at lower scales. Streaming separate models resolves these issues but requires more bandwidth. Neither option is ideal for battery life in mobile devices. Our multi-scale model alleviates the aliasing and flickering artifacts without incurring the drawbacks of storing and transmitting multiple models.

\begin{table}[!htbp]
  \caption{Training Ablation Average Rendering Quality Results.}
  \newcommand{\quantitativeQualityTabScale}{0.9}
  \label{tab:ablation_quality}
  \centering
  \begin{subfigure}[h]{\linewidth}
      \centering
      \scalebox{\quantitativeQualityTabScale}{
      \begin{tabular}{lcccc}
          \toprule
          Ablation & LOD 1 & LOD 2 & LOD 3 & LOD 4 \\ 
          \midrule
Progressive Training & 26.30 & 30.33 & 29.08 & 28.10 \\
108 Training Views   & 29.04 & 29.63 & 28.93 & 27.98 \\
Ours                 & 29.37 & 29.88 & 29.01 & 28.12 \\
          \bottomrule
      \end{tabular}
      }
      \caption{PSNR (dB) at 1/8, 1/4, 1/2, and 1/1 scale.}
      \label{fig:ablation_quality_psnr}
  \end{subfigure}\\
  \vspace{2mm}
  \begin{subfigure}[h]{\linewidth}
      \centering
      \scalebox{\quantitativeQualityTabScale}{
      \begin{tabular}{lcccc}
          \toprule
           Model & LOD 1 & LOD 2 & LOD 3 & LOD 4 \\ 
          \midrule
Progressive Training & 0.7947 & 0.8719 & 0.8447 & 0.8390 \\
108 Training Views   & 0.8765 & 0.8771 & 0.8604 & 0.8566 \\
Ours                 & 0.8834 & 0.8819 & 0.8626 & 0.8570 \\
           \bottomrule
      \end{tabular}
      }
      \caption{SSIM at 1/8, 1/4, 1/2, and 1/1 scale.}
      \label{fig:ablation_quality_ssim}
  \end{subfigure}
\end{table}

\subsection{Training Ablation}
To evaluate our training strategy, we perform two ablation experiments.
First, we compare our strategy to a coarse-to-fine progressive training strategy where lower levels of detail are trained and then frozen as higher levels are trained.
Progressive training takes on average $24.20$ hours to train.
Second, we use half the number of training views to evaluate how the quality degrades with fewer training views.
Both experiments use our progressive multi-scale model architecture.
Our results are shown in~\autoref{tab:ablation_quality} with additional details in the supplementary material.

\begin{figure}[!htbp]
    \centering
    \includegraphics[width=\linewidth]{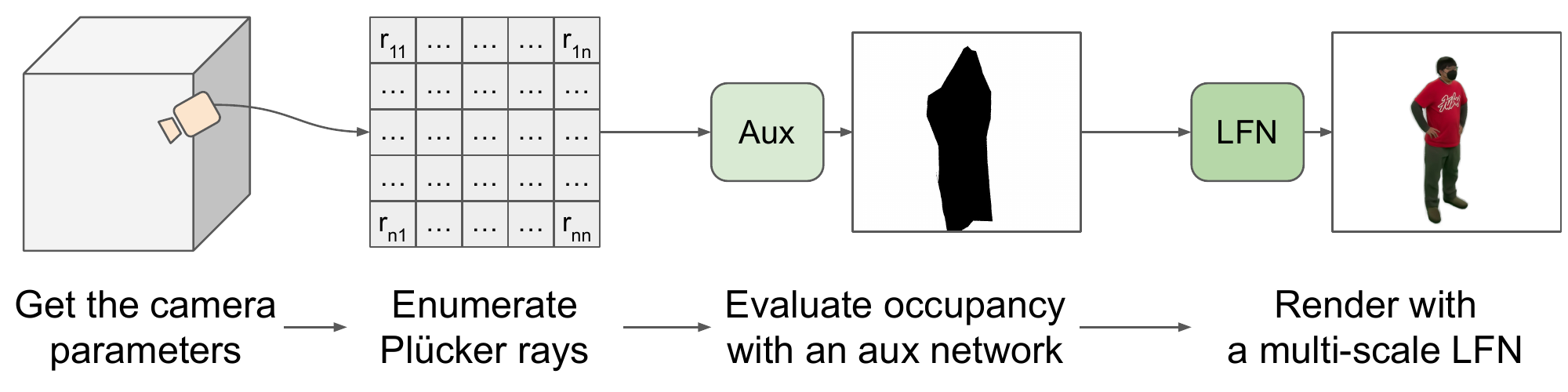}
    \caption{Overview of our rendering pipeline which utilizes an auxiliary network to skip evaluation of empty rays.}
    \label{fig:rendering_pipeline}
\end{figure}

\begin{table}[!htbp]
  \caption{Average Model Rendering Performance for our Multi-scale LFN (milliseconds per frame).}
  \label{tab:performance_table_v2}
  \begin{subfigure}[t]{\linewidth}
      \centering
      \begin{tabular}{cccc}
        \hline
         LOD 1 & LOD 2 & LOD 3 & LOD 4\\
        \hline 
3.7 & 3.9 & 4.7 & 5.9\\
        \hline
      \end{tabular}
      \caption{Rendering each LOD at 1/8 scale.}
      \label{tab:performance_table_v2_8thscale}
  \end{subfigure}\\%
  \begin{subfigure}[!ht]{\linewidth}
      \centering
      \begin{tabular}{cccc}
        \hline
        LOD 1 & LOD 2 & LOD 3 & LOD 4\\
        \hline
4 & 11 & 58 & 305\\
        \hline
      \end{tabular}
      \caption{Rendering at 1/8, 1/4, 1/2, and 1/1 scale.}
      \label{tab:performance_table_v2_multiscale}
  \end{subfigure}
\end{table}

\subsection{Level of Detail Rendering Speedup}
We evaluate the rendering speed of our progressive model at different levels of detail to determine the observable speedup from rendering with the appropriate level of detail. For optimal rendering performance, we render using half-precision floating-point values and skip empty rays using an auxiliary neural network encoding only the ray occupancy as shown in \autoref{fig:rendering_pipeline}. Our auxiliary network is made up of three layers with a width of 16 neurons per hidden layer. Evaluation is performed by rendering rays from all training poses with intrinsics scaled to the corresponding resolution as shown in \autoref{tab:model_parameters}. 
Average rendering time results are shown in \autoref{tab:performance_table_v2}. 
Rendering times for lower LODs (1/8 scale) from our multi-scale network are reduced from $\sim5.9$ to $\sim3.7$ msec.
\section{Discussion}
The primary contribution of our paper is the identification of a representation more suitable for on-demand streaming. 
This is achieved by enhancing light field networks, which support real-time rendering, to progressively support multiple levels of detail at different resolutions.
Our representation inherits some of the limitations of LFNs such as long training times and worse view-synthesis quality compared to NeRF-based models.
Since our method does not support continuous LODs, we require the user to decide the LODs ahead of time. For intermediate resolutions, rendering to the nearest LOD is sufficient to avoid aliasing and flickering.
While our light field datasets could also be rendered with classical techniques~\cite{levoy1996lightfieldrendering,xiao2014cvpr,peter2001wavelet}, doing so over the internet would not be as bandwidth-efficient.
Future work may combine our multi-scale light field networks with PRIF~\cite{Feng2022PRIF} to encode color and geometry simultaneously or with VIINTER~\cite{Feng2022Viinter} for dynamic light fields.
\section{Conclusion}
In this paper, we propose a method to encode multi-scale light field networks targeted for streaming.
Multi-scale LFNs encode multiple levels of details at different scales to reduce aliasing and flickering at lower resolutions.
Our multi-scale LFN features a progressive representation that encodes all levels of details into a single network using subsets of network weights. 
With our method, light field networks can be progressively downloaded and rendered based on the desired level of detail for real-time streaming of 6DOF content.
We hope progressive multi-scale LFNs will help improve the real-time streaming of photorealistic characters and objects to real-time 3D desktop, AR, and VR applications~\cite{li2020meteovis,du2019geollery}.
{\small\paragraph*{Acknowledgments}
We would like to thank Jon Heagerty, Sida Li, Barbara Brawn-Cinani, and Maria Herd for developing our light field datasets as well as the anonymous reviewers for the valuable comments on the manuscript.
This work has been supported in part by the NSF Grants 18-23321 and 21-37229, and the State of Maryland’s MPower initiative. 
Any opinions, findings, conclusions, or recommendations expressed in this article are those of the authors and do not necessarily reflect the views of the research sponsors.
}

{\small
\bibliographystyle{ieee_fullname}
\bibliography{egbib}
}

\end{document}